\documentclass[letterpaper]{article} % DO NOT CHANGE THIS
\usepackage{aaai24}  % DO NOT CHANGE THIS
\usepackage{times}  % DO NOT CHANGE THIS
\usepackage{helvet}  % DO NOT CHANGE THIS
\usepackage{courier}  % DO NOT CHANGE THIS
\usepackage[hyphens]{url}  % DO NOT CHANGE THIS
\usepackage{graphicx} % DO NOT CHANGE THIS
\usepackage{comment}
\usepackage{appendix}
\usepackage{natbib}  % DO NOT CHANGE THIS AND DO NOT ADD ANY OPTIONS TO IT
\usepackage{caption} % DO NOT CHANGE THIS AND DO NOT ADD ANY OPTIONS TO IT
\usepackage{algorithm}
\usepackage{algorithmic}
\usepackage{multirow}
\usepackage{multicol}
\usepackage{amsmath,amsfonts,amssymb,amsthm, bm}
\usepackage{xcolor}
\newcommand\mdoublepluss{\tiny \mathbin{+~ \mkern-10mu +}}

\usepackage{newfloat}
\usepackage{listings}
\DeclareCaptionStyle{ruled}{labelfont=normalfont,labelsep=colon,strut=off} % DO NOT CHANGE THIS
\floatstyle{ruled}
\newfloat{listing}{tb}{lst}{}
\floatname{listing}{Listing}
\title{Spatio-Temporal driven Attention Graph Neural Network with Block Adjacency matrix (STAG-NN-BA) for Remote Land-use Change Detection}
\author {
    Usman Nazir\textsuperscript{\rm 1},
    Wadood Islam\textsuperscript{\rm 1},
    Sara Khalid\textsuperscript{\rm 2},
    Murtaza Taj\textsuperscript{\rm 1}
}
\affiliations {
    \textsuperscript{\rm 1} Lahore Uniersity of Management Sciences\\
    \textsuperscript{\rm 2} University of Oxford\\
    \{usman.nazir,muhammad\_islam, murtaza.taj\}@lums.edu.pk \\
    sara.khalid@ndorms.ox.ac.uk
}

\usepackage{bibentry}
\begin{document}

\maketitle

\begin{abstract}
Land-use monitoring is fundamental for spatial planning, particularly in view of compound impacts of growing global populations and climate change. Despite existing applications of deep learning in land use monitoring, standard convolutional kernels in deep neural networks limit the applications of these networks to the Euclidean domain only. Considering the geodesic nature of the measurement of the earth's surface, remote sensing is one such area that can benefit from non-Euclidean and spherical domains. For this purpose, we designed a novel Graph Neural Network architecture for spatial and spatio-temporal classification using satellite imagery to acquire insights into socio-economic indicators. We propose a hybrid attention method to learn the relative importance of irregular neighbors in remote sensing data. Instead of classifying each pixel, we propose a method based on Simple Linear Iterative Clustering (SLIC) image segmentation and Graph Attention Network. The superpixels obtained from SLIC become the nodes of our Graph Convolution Network (GCN). A region adjacency graph (RAG) is then constructed where each superpixel is connected to every other adjacent superpixel in the image, enabling information to propagate globally. Finally, we propose a Spatially driven Attention Graph Neural Network (SAG-NN) to classify each RAG. We also propose an extension to our SAG-NN for spatio-temporal data. Unlike regular grids of pixels in images, superpixels are irregular in nature and cannot be used to create spatio-temporal graphs. We introduce temporal bias by combining unconnected RAGs from each image into one supergraph. This is achieved by introducing block adjacency matrices resulting in novel Spatio-Temporal driven Attention Graph Neural Network with Block Adjacency matrix (STAG-NN-BA). We evaluated the proposed methods on two remote sensing datasets namely Asia14 and C2D2. In comparison with both non-graph and graph-based approaches our SAG-NN and STAG-NN-BA achieved superior accuracy on both datasets while incurring less computation cost. The code~\footnote{\url{https://github.com/usmanweb/Codes}} and dataset is publicly available.
\end{abstract}

\section{Climate Impact Pathway}
Throughout history there have been changes in land-use in part due to development and farming practices. Combined with population growth these activities have had a lasting impact on the environment. The United Nations Development Programme (UNDP) estimated that the in the period between 1950 and 2015, city-dwelling population increased from 54.6\% to 78.3\% \cite{un2016world}. With unprecedented population growth globally, humans increasingly continue to claim  forests to develop cities, industries, and farms to meet the increasing demands of living spaces and and food supply chains. Forest and green spaces are consequently shrinking, in turn exacerbating global warming and related impacts such as wildfires. In order to mitigate impacts and develop adaptation plans, monitoring and modelling land-use is of utmost importance.

The proposed approach, aimed at harnessing the potential of spatial and spatio-temporal data, is not limited to classification tasks but extends its utility to detect critical transitions in land-use. Specifically, transitions between classes such as construction and destruction, cultivation and decultivation - representing fundamental human activities with historical significance - can be effectively identified. 
Given the importance of monitoring land-use, afforestation and deforestation activity in measuring emissions and decarbonisation efforts, the proposed approach can contribute to enhanced understanding and actionable insights in addressing climate change impacts and mitigation strategies.   
%%Wadood

\section{Introduction}
 
In order to inform future spatial planning, it is crucial to study land-use leveraging spatial data from the past and present \cite{dadras2015spatio}. Changes in land use over time can be gauged by analyzing satellite imagery via spatio-temporal analysis. Machine learning derived modelling of satellite imagery can help to reliably account for the destruction brought about by conflict, deforestation, and natural disasters. This requires approaches that can intelligently recognize and categorize geographical changes in land-use or land-cover. 

\begin{figure*}
    \centering
    \scalebox{0.66}{
    \begin{tabular}{cc}
     \includegraphics[scale=0.2]{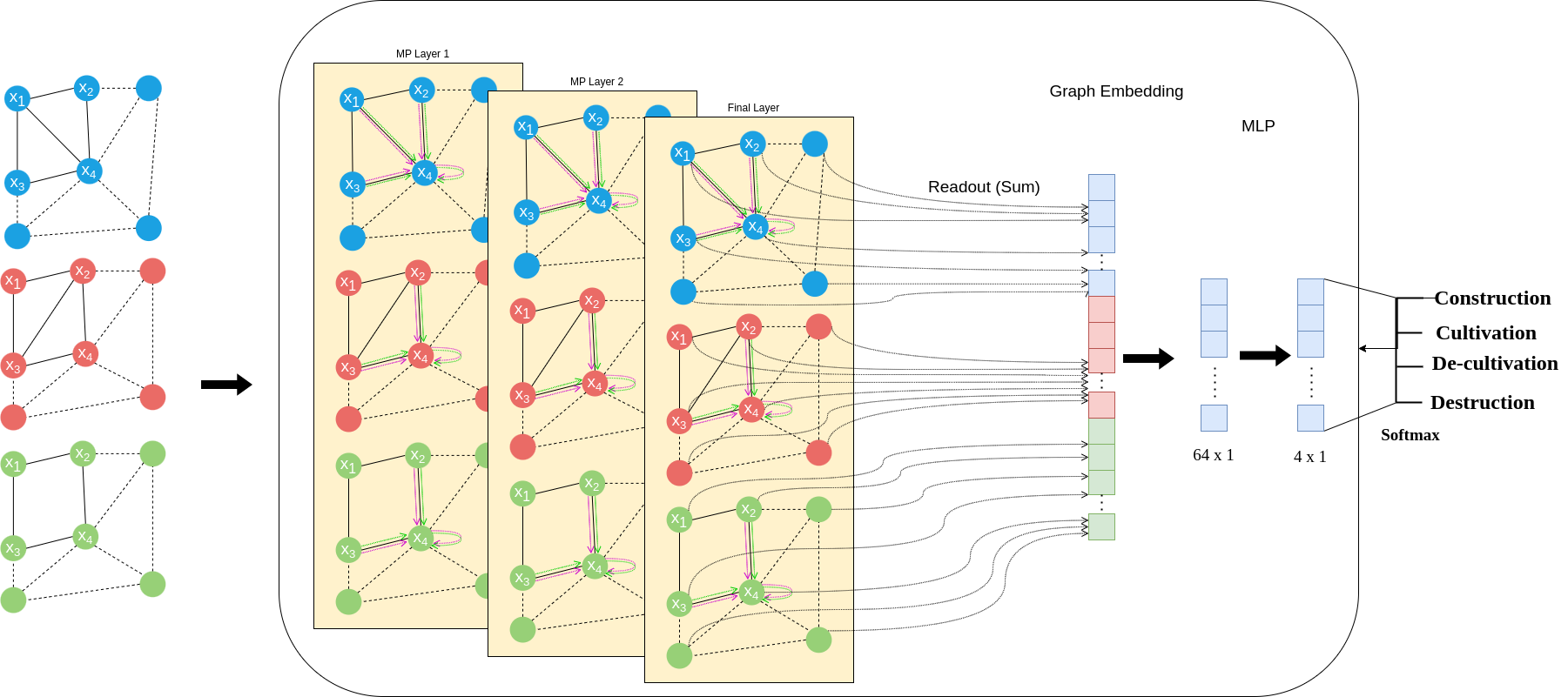} & \includegraphics[scale=0.2]{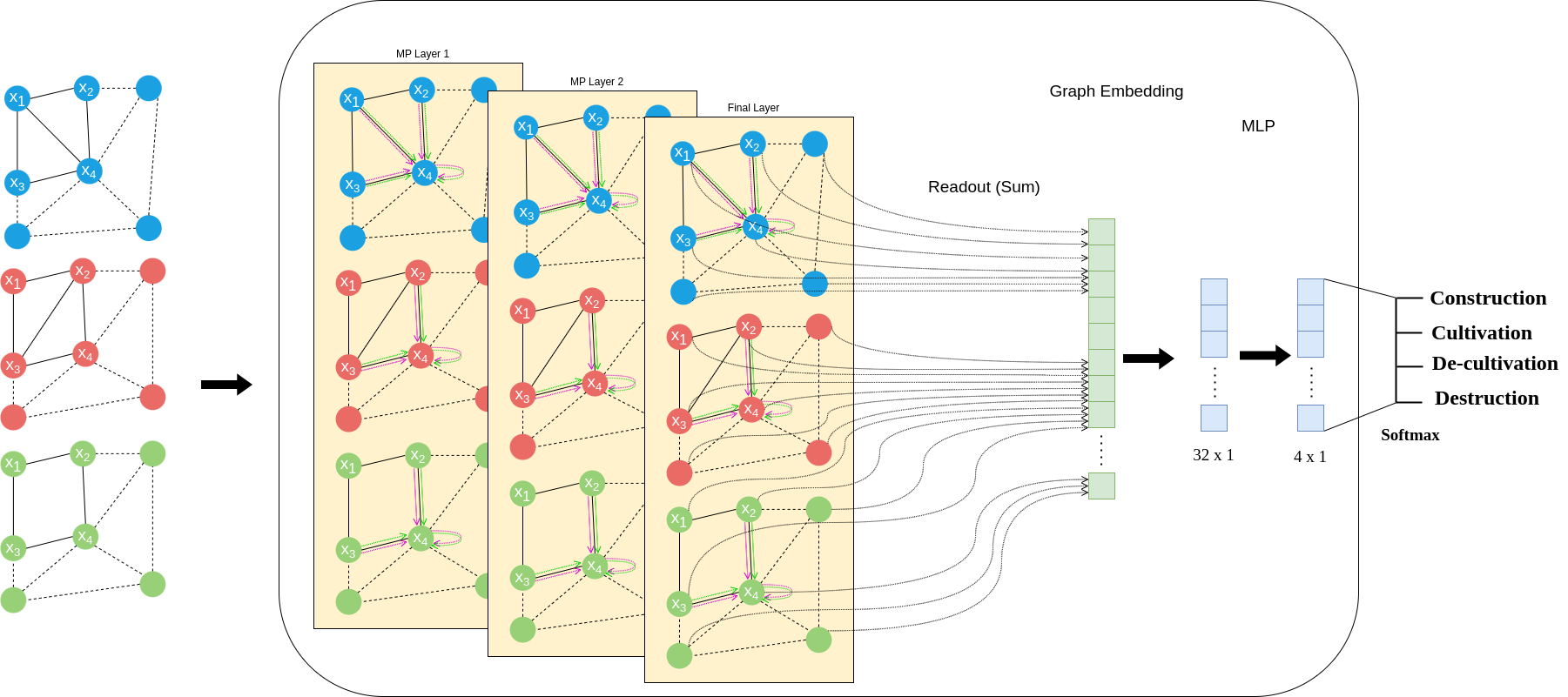} \\
     STAG-NN-BA-GCP & STAG-NN-BA-GSP \\
    \end{tabular}}
    \caption{Spatio-Temporal driven Attention Graph Neural Network with Block Adjacency matrix (STAG-NN-BA). }
    \label{fig:stag-nn-ba12}
\end{figure*}
Deep learning, particularly CNNs have in the recent past revolutionized many machine learning tasks. Examples include image classification~\cite{krizhevsky2017imagenet, li2019deep, zhang2019medical}, video processing~\cite{sharma2021video, sreenu2019intelligent}, speech recognition~\cite{laux2023two, zhang2022wenet}, and natural language processing~\cite{zhu2022knowledge, lucic2022towards}. These applications are usually characterized by data drawn from Euclidean space. However, measurements over the surface of the earth are inherently non-Euclidean in nature due to its irregular and changing shape, height due to mountains, and depth due to deep ocean trenches. Data from such non-Euclidean space can be represented as graphs~\cite{kipf2016semi, bliss2013confronting, velickovic2017graph} so as to capture the complex relationships and inter-dependency between objects. Recently, many studies on extending deep learning approaches for graph data have emerged~\cite{henaff2015deep,defferrard2016convolutional,jain2016structural,kipf2016semi,wang2018deep,satorras2018few,narasimhan2018out,hu2018relation,gu2018learning,wang2018zero,lee2018multi,qi2018learning,marino2016more,kampffmeyer2019rethinking,edwards2016graph,liu2020cnn,fey2019fast,wan2019multiscale,qi2018stagnet,zhou2019relation,zhang2020spatio,tompson2014joint}. For instance, graph neural networks (GNNs) have been increasingly used for applications such as molecule and social network classification \cite{knyazev2018amer} and generation \cite{simonovsky2017dynamic}, 3D Mesh classification and correspondence \cite{fey2018splinecnn}, modeling behavior of dynamic interacting objects \cite{kipf2018neural}, program synthesis \cite{allamanis2017learning}, reinforcement learning tasks \cite{bapst2019structured} and other domains.
\begin{comment}
\begin{figure}
    \centering
    \includegraphics[scale=0.5]{newimages/STAG-NN-BA.pdf}
    \caption{Overview of hybrid attention mechanism of GATv3. In the dashed rectangle we compute similarity score between neighboring nodes and take dot-product with the attention mechanism of original GAT~\cite{velivckovic2017graph}.}
    \label{proposed}
\end{figure}
\end{comment}

While the utility of graph neural networks for emerging applications is promising, the complexity of graph data imposes significant challenges on many existing machine learning algorithms. For instance, in the field of image processing, the use of Graph Convolutional Networks (GCN) is still limited to a few examples ~\cite{kampffmeyer2019rethinking,wang2018zero,lee2018multi}. Through carefully hand-crafted graph construction methods or other supervised approaches, images can be converted to structured graphs capable of processing by GCNs. In these GNNs, each pixel of an image is considered as a graph node~\cite{edwards2016graph} which is cumbersome and in many cases unnecessary. Instead of learning from raw image pixels, the use of 'superpixels' addresses this concern \cite{liang2016semantic,knyazev2019image} and helps in reducing the graph size and thereby the computational complexity. The applications of Superpixels include saliency estimation~\cite{zhu2014saliency}, optical flow estimation~\cite{sevilla2016optical}, object detection~\cite{yan2015object}, semantic segmentation~\cite{gadde2016superpixel}, reduce input for subsequent algorithms~\cite{fey2019fast} and explainable AI~\cite{ribeiro2016should}.

In this paper, we propose a hybrid attention method to incorporate these relational inductive biases in remote sensing data. Instead of classifying each pixel, we propose a method based on Simple Linear Iterative Clustering (SLIC) image segmentation and Graph Attention Network: GAT~\cite{velickovic2017graph} to detect socio-economic indicators from remote sensing data. We first over-segment the image into superpixels. These superpixels become the nodes of our Graph Convolution Network (GCN). We then construct a region adjacency graph (RAG) where each superpixel is connected to every other adjacent superpixel in the image, enabling information to propagate globally. Finally, we classify each RAG via Spatially driven Attention Graph Neural Network (SAG-NN). We also propose an extension to our SAG-NN for spatio-temporal data named as Spatio-temporal Attention driven GNN (STAG-NN). Unlike, pixels or objects, superpixels are prone to change over time, to address this problem we propose a STAG-NN with Block diagonal Adjacency matrix (STAG-NN-BA) which enables us to incorporate both the spatial as well as temporal information in a single time-varying graph. The primary novelty of this paper is the SAG-NN and STAG-NN-BA architectures for the prediction of spatio-temporal transition classes (such as construction, destruction, cultivation, and harvesting) from remote sensing data. We demonstrate that this approach incurs a smaller computational cost compared with other deep learning methods. The details of our proposed approach, which is derived from vanilla GAT~\cite{velickovic2017graph}, are presented in Section~\ref{sec:proposed}.

In this paper, we propose a unified framework allowing to generalize geometric deep learning to remote sensing data and learn spatial and spatio-temporal features using superpixels. We improve the GAT scoring function to overcome the following shortcomings in GATv1~\cite{velickovic2017graph} and GATv2~\cite{brody2021attentive}: 1) In GATv1, the learned layers $\mathbf{W}$ and $a$ are applied consecutively, and thus can be collapsed into the single linear layer. 2) GATv2~\cite{brody2021attentive} performs best for a complete bipartite graph. We improved the graph attention scoring function by introducing the relational inductive bias in data using neighborhood features aggregation as well as the ranking of attended nodes. Our proposed approach achieves higher accuracy with less computing cost than state-of-the-art graph neural network architectures.

\section{Challenges}\label{cdpre}

\subsection{Heterogeneity in Remote Sensing Data} \label{sec:chsol}
While considering a large geographic area, several inherent complexities in satellite imagery make automated detection of change in land-use a challenging task. This includes, but is not limited to, i)  variations in imaging sensors, ii) differences in construction design across the countries, iii) dynamic surroundings and iv) variations in luminosity, seasonal changes, and pollution levels, etc. 

The heterogeneity in types of land surface cover, in particular, poses a major challenge for the task of spatial and spatio-temporal analysis. High resolution satellite imagery has drawn much attention from the scientific community due to the fine spatial details of land surface covers. Pixel-based classification methods are hardly applicable for high-resolution remote sensing images due to the high interior heterogeneity of land surface covers. The separation between spectral signatures of different land surface covers is more difficult due to the abundant details in pixel-based classification~\cite{zhang2019mapping}. To deal with this challenge, we are using superpixel-based classification which reduces the redundancy of the spatial features of different ground objects. Details of other challenges can be found in this paper~\cite{nazir2020kiln}.

\subsection{Representation of Images as Graphs}
GNNs on images are characterized by unique challenges with respect to their implementation. Most of the graph neural frameworks~ \cite{defferrard2016convolutional,edwards2016graph,liu2020cnn} are designed for dense representations such as pixel-based graphs. However, pixel based representation results in a large number of nodes which increases both the compute as well as memory costs. Since adjacent pixels are known to have similar information except at object boundaries, pixel based representation is not only cumbersome, but it is also highly redundant. To address this concern superpixel and object-based graphs have been extensively used in the literature~\cite{liang2016semantic,knyazev2019image,fey2019fast,liu2020cnn,wan2019multiscale,jain2016structural,qi2018stagnet,zhou2019relation,zhang2020spatio,jiang2013hallucinated,tompson2014joint}. For subsequent processing, superpixels have been widely used as an effective way to reduce the number of image primitives. 

The literature includes numerous methods for determining a superpixel based representation from an image, each with different strengths and weaknesses. Recently, many DNN-based methods to identify superpixels have been proposed~\cite{yang2020superpixel,jampani2018superpixel}. But the most popular of practices in the GNN literature (on account of generally good results and low compute complexity) are SLIC~\cite{achanta2012slic}, Quickshift~\cite{vedaldi2008quick} and Felzenszwalb~\cite{felzenszwalb2004efficient}. Details of these methods are presented in the following subsections.

%One can achieve effective parallelization on irregular sparse graphs (e.g., superpixel-based graphs) by using fast localized spectral filtering~\cite{defferrard2016convolutional} or hierarchichal multi graph spatial convolutional neural networks \cite{knyazev2019image}. 

\subsubsection{SLIC}
The SLIC (simple linear iterative clustering)~\cite{achanta2012slic} algorithm  performs an iterative clustering approach in the 5D space of color information and image location. The algorithm quickly gained momentum and is now widely used due to its speed, storage efficiency, and successful segmentation in terms of color boundaries. However, the main limitation of SLIC is that it often captures background pixels as shown in Fig. ~\ref{fig:superpixels} -- Column 1, and therefore does not significantly help in data reduction for the graph generation. However it performs better in capturing built-up and grassy land from satellite imagery as shown in Fig.~\ref{fig:sRAG} -- Column 2.

\subsubsection{Quickshift}\label{qs}
Quickshift~\cite{vedaldi2008quick} is a relatively recent 2D algorithm that is based on an approximation of kernelized mean-shift~\cite{comaniciu2002mean}. It segments an image based on the three parameters: $\epsilon$ for the standard deviation of the Gaussian function, $\alpha$ for the weighting of the color term, and $S$ to limit the calculating a window size of $S \times S$. Therefore, it belongs to the family of local mode-seeking algorithms and is applied to the $5$D space consisting of color information and image location. One of the benefits of Quickshift is that it actually computes a hierarchical segmentation on multiple scales simultaneously. As shown in Fig.~\ref{fig:superpixels} -- Column 2, it does not capture background pixels and also reduces $30\%$ of input data for the graph generation. But it cannot segment built-up and grassy areas perfectly as shown in Fig.~\ref{fig:sRAG} -- Column 3.

\subsubsection{Felzenszwalb}
\begin{figure}[h!]
\centering
\scalebox{1}{
	\begin{tabular}{ccc}
	    \includegraphics[width=.28\columnwidth]{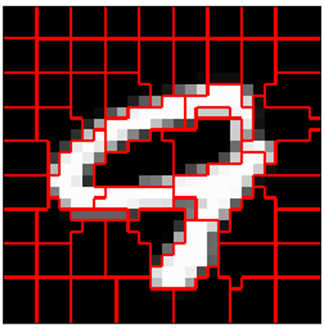} &
	    \includegraphics[width=.28\columnwidth]{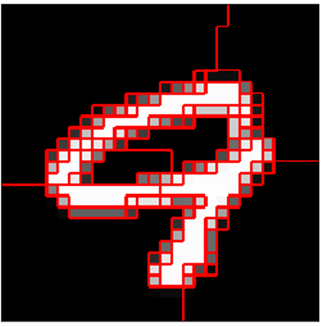} &
        \includegraphics[width=.28\columnwidth]{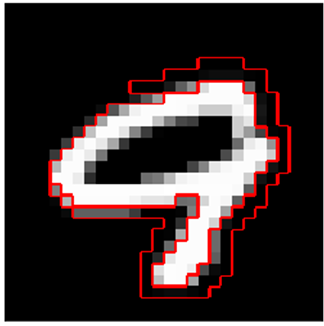}\\
        SLIC & Quickshift & Felzenszwalb  \\
		\end{tabular}}
	\caption{Superpixel segmentation techniques on MNIST digit: 9.}
	\label{fig:superpixels}
\end{figure}

\begin{figure}[h!]
\centering
\scalebox{1}{
	\begin{tabular}{ccc}
        \includegraphics[width=.28\columnwidth]{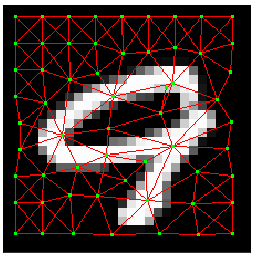} &
	    \includegraphics[width=.28\columnwidth]{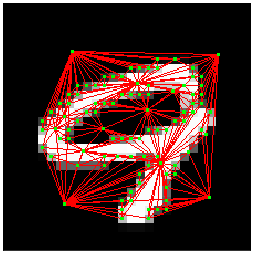} &
        \includegraphics[width=.28\columnwidth]{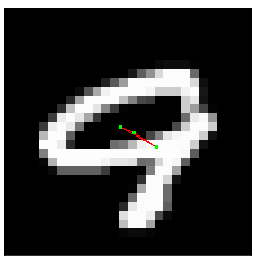}
        \\
		\end{tabular}}
	\caption{Region Adjacency Graphs (RAG) generation from SLIC, Quickshift and Felzenszwalb superpixels respectively.}
	\label{fig:RAGs}
\end{figure}

\begin{figure}[h!]
\centering
\scalebox{1}{
	\begin{tabular}{ccc}
	    \includegraphics[width=0.28\columnwidth]{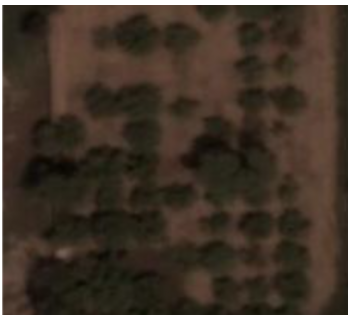}&\includegraphics[width=0.28\columnwidth]{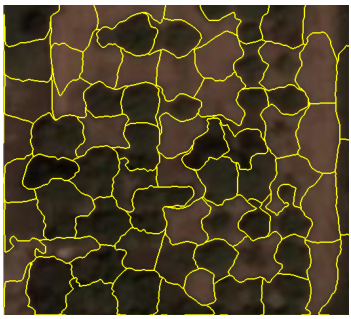}&\includegraphics[width=0.28\columnwidth]{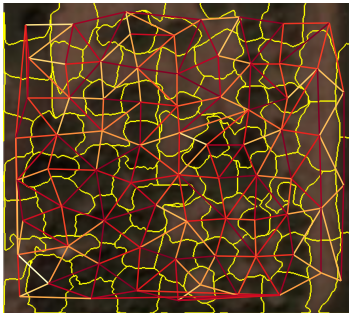}\\
	   \includegraphics[width=0.28\columnwidth]{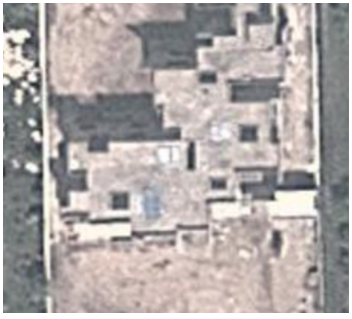}&\includegraphics[width=0.28\columnwidth]{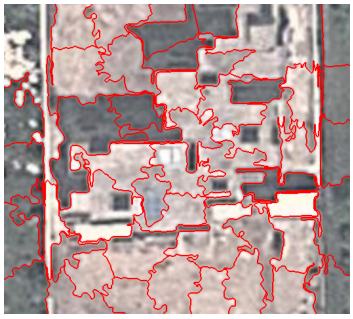}&\includegraphics[width=0.28\columnwidth]{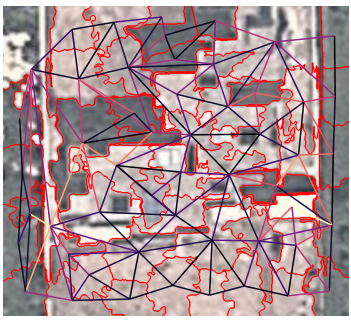}\\
	   Satellite Image & SLIC superpixels & Adjacency matrix \\
		\end{tabular}}
	\caption{RAG generation from SLIC superpixels on image from Asia14 dataset (Satellite images courtesy Google Earth).}
	\label{fig:sRAG}
\end{figure}

This fast 2D image segmentation algorithm, proposed in ~\cite{felzenszwalb2004efficient}, has a single scale parameter that influences the segment size. The actual size and number of segments can vary greatly, depending on local contrast. This segmentation appeared to be less suitable in tests on a series of images, as its parameters require a special adjustment, and consequently, a static choice of this parameter leads to unusable results. As shown in Fig.~\ref{fig:superpixels} -- Column 3 and Fig.~\ref{fig:sRAG} -- Column 1, it only captures the pixels corresponding to the region of interest pixels but performs poorly in graph generation procedure as shown in Fig. \ref{fig:RAGs} - Column 3.

\begin{figure}[h!]
\centering
\scalebox{0.9}{
	\begin{tabular}{cccc}
	    \includegraphics[width=0.25\columnwidth]{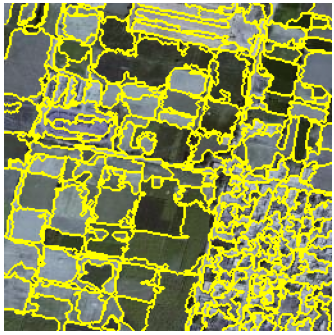}&
     \includegraphics[width=0.23\columnwidth]{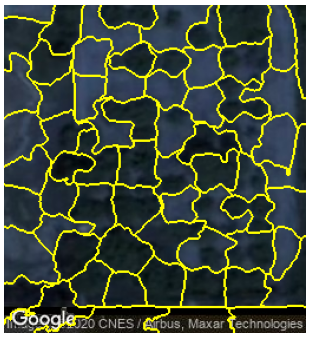}&
     \includegraphics[width=0.23\columnwidth]{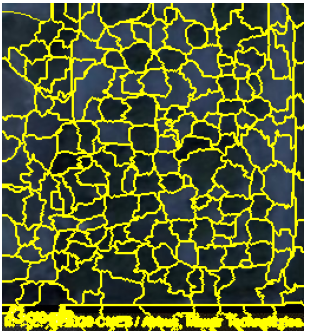}&
     \includegraphics[width=0.23\columnwidth]{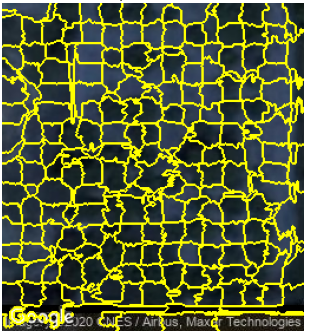}\\
	    Felzenszwalbs's & SLIC & Quickshift & C. Watershed\\
		\end{tabular}}
	\caption{Superpixel segmentation techniques on image from Asia14 dataset. Felzenszwalbs's method and quickshift cannot segment perfectly built-up and barren land due to inherent complexities in satellite imagery. On the other hand, compact watershed poorly performed on grassy land. While SLIC works perfectly on satellite imagery.}
	\label{fig:ssuperpixels}
\end{figure}

Instead of grid-based placement as in images, superpixels usually result in irregular representation depending upon image content. Such irregular representation restricts the construction of graph on spatio-temporal data. This work has addressed this issue by proposing STAG-NN-BA which resolves the issue via a block adjacency matrix.

\section{Proposed Methodology}
\label{sec:proposed}
The proposed methodology consist of following major steps: 
\begin{itemize}
    \item Generate a superpixel representation of the input images.
    \item Create a region adjacency graph (RAG) from the superpixel representation, by connecting  neighbouring superpixels.
    \item Spatial Attention Graph Neural Network (SAG-NN) from region adjacency graph (RAG) for spatial classification.
    \item Spatio-temporal driven Graph Attention Neural Network with Block Adjacency matrix (STAG-NN-BA) for classification of transitions or changes in land-use over time.
\end{itemize}

The following subsections discuss the proposed architecture in detail.

\subsection{Superpixel Segmentation}
When we apply segmenation techniques on satellite imagery, SLIC~\cite{achanta2012slic} performs better as compared to Quickshift~\cite{vedaldi2008quick},  Felzenszwalb~\cite{felzenszwalb2004efficient} and Compact watershed~\cite{neubert2014compact}. As shown in Fig.~\ref{fig:ssuperpixels} -- Column 2, SLIC captures the color boundaries, and segments perfectly the built-up area and agricultural land. It is more stable for satellite imagery as compared to other segmentation techniques. The superpixel segmentation technique using SLIC~\cite{achanta2012slic} provides an elegant way to divide the satellite image into homogeneous regions as shown in Fig.~\ref{fig:ssuperpixels}. We set the number of segments to $75$ and compactness to $10$. This resulted in approximately $75$ superpixels per image and subsequently a graph of $75$ nodes instead of $65536$ nodes in case of using raw pixel values of remote sensing imagery.

\subsection{Graph generation from superpixels}
%This subsection discusses two methods used for generating graphs from superpixels. These are presented in the following.

%\subsubsection{Region Adjacency Graph (RAG)}\label{rag}
After using a superpixel segmentation technique, a Region Adjacency Graph (RAG) is generated by treating each superpixel as a node and adding edges between all directly adjacent superpixels. Unlike MoNet~\cite{monti2017geometric}, which use K-Nearest Neighbours to form a connection between nodes, in our graph $G$ we formed connections based on immediate adjacency only. Thus ours is a more compact graph while the information from neighbours of neighbours can still be incorporated in this case by using K-hop messaging passing. Each graph node can have associated features, providing aggregate information based on the characteristics of the superpixel itself. The regions obtained in the segmentation stage are represented as vertices $V$ and relations between neighboring regions are represented as edges $E$. The search for the most similar pair of regions is repeated several times per iteration and every search requires $\mathcal{O}(N)$ region similarity computations. The graph is utilized so that the search is limited only to the regions that are directly connected by the graph structure.

\begin{figure}[h!]
\centering
\includegraphics[width=7cm]{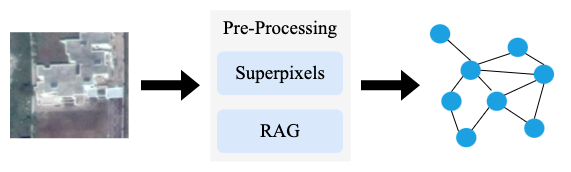}
 \caption{RAG Generation from a single geospatial image.}
\label{fig:RAG}
\end{figure}
\begin{figure}[h]
\centering
\includegraphics[width=7cm]{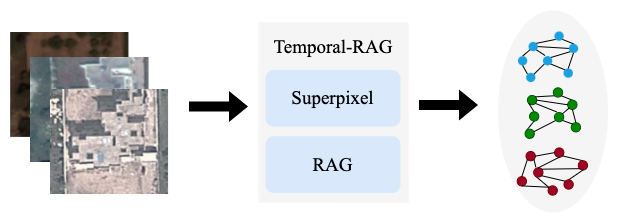}
 \caption{Generation of Temporal RAG from geospatial images of same geolocation from multiple years.}
\label{fig:TRAG}
\end{figure}

\subsection{Spatial Attention Graph Neural Network (SAG-NN)}
We will start by describing a single message passing layer, as the sole layer utilized throughout all of the GCN~\cite{kipf2016semi} and GAT~\cite{velickovic2017graph} architectures.

Consider a graph $G(V, E)$, where $V$ is set of $n$ nodes and $E$ is the set of $m$ vertices. $G$ is specified as a set of nodes' initial embeddings (input features): $(\overrightarrow{x_1}, \overrightarrow{x_2},\dots,\overrightarrow{x_n})$, and an adjacency matrix ${\mathbf {ADJ}}$, such that ${\mathbf {ADJ}}_{i,j} = 1$ if $i$ and $j$ are connected, and $0$ otherwise. Consider node $i$'s initial embedding (for step $k=0$) is: 
\begin{equation}
    \overrightarrow h_i^{(0)} = \overrightarrow x_i, \forall i \in V 
\end{equation}

A graph convolutional layer at step $k = 1, 2, \dots, K$ then computes a set of new node features (${\overrightarrow{h_1}^{k}}, {\overrightarrow{h_2}^{k}},\dots,{\overrightarrow{h_n}^{k}}$), based on the input features as well as the graph structure. Every graph convolutional layer starts off with a shared feature transformation specified by a weight matrix $\mathbf W$. 
%This tranforms the feature vectors into $\overrightarrow{h_i}=\mathbf W\overrightarrow{x_i}$. After this, the vectors $\overrightarrow{g_i}$ are typically recombined in some way at each node.

In general, to satisfy the localization property, we will define a graph convolutional operator as an aggregation of features across neighbourhoods; defining $\mathcal{N}_i$ as the neighbourhood of node $i$ (typically consisting of all first-order neighbours of $i$, including $i$ itself), we can define the output features of node $i$ as
%\sigma \Bigg(\mathbf{W} {[\overrightarrow {h_i} || \overrightarrow {h_j}]} \Bigg)
\begin{equation}
  \overrightarrow{{h}}_i^{(k)} = f^{(k)} \Bigg( \mathbf{W}^{(k)}\cdot \Bigg[\sum_{j\in{\mathcal{N}_i}} C^{(k)}\overrightarrow{{h}}_j^{(k-1)} + C^{(k)} \overrightarrow h_i^{(k-1)} \Bigg] \Bigg )
  %{\color{red}\Bigg\vert~\cite{kipf2016semi}} 
\end{equation}  
where $\forall i\in V$ and $f^{(k)}$ is an activation function. Each neighbour can be assigned different importance as:

\begin{equation}\label{vanillaGAT}
    \overrightarrow h_i^{(k)} = f^{(k)} \Bigg(\mathbf{W}^{(k)} . \Bigg[\sum_{j \in \mathcal{N}_i} \alpha_{ij}^{(k-1)} \overrightarrow h_j^{(k-1)} + \alpha_{ii}^{(k-1)} \overrightarrow h_i^{(k-1)} \Bigg] \Bigg )
    %{\color{red}\Bigg\vert~\cite{velickovic2017graph}}
\end{equation}
where $\forall i\in V$ and $\sum_{j \in \mathcal{N}_i}(.)$ is the weighted mean of i's neighbour's embedding at step $k-1$ and the attention weights $\alpha^{(k)}$ are generated by an attention mechanism $\mathbf A^{(k)}$, normalized such that the sum over all neighbours of each node i is 1:
\begin{equation}\label{vanillaGATAttention}
\alpha_{ij}^{(k)} = \frac{\mathbf{A}^{(k)}(\overrightarrow{h}_i^{(k)}, \overrightarrow{h}_j^{(k)})}{\sum_{w \in \mathcal{N}_i} \mathbf{A}^{(k)}(\overrightarrow{h}_i^{(k)}, \overrightarrow{h}_w^{(k)})},~\forall (i, j)\in E
\end{equation}

%where $\eta(\overrightarrow {h_i}, \overrightarrow {h_j})$ specifies the weighting factor (importance) of nodes $j$'s features to node $i$. More prior work defines weighting factor explicitly either based on the structural properties of the graph or as a learnable weight; this requires compromising at least one other desirable property.
%\textbf{Add kipf and maxwell papers}

In standard GAT (see eq.~\ref{vanillaGAT} \& \ref{vanillaGATAttention}) $\alpha_{ij}$ is implicitly defined, employing self-attention over the node features to do so. This choice was not without motivation, as self-attention has previously been shown to be self-sufficient for state-of-the-art-level results on machine translation, as demonstrated by the Transformer architecture~\cite{vaswani2017attention}.

Generally, we let $\alpha_{ij}$ be computed as a byproduct of an attentional mechanism, $a: \mathcal{R}^N \times \mathcal{R}^N \longrightarrow \mathcal{R}$ which computes normalized coefficients $\alpha_{ij}$ across pairs of nodes $i,j$, based on their features (see eq.~\ref{vanillaGATAttention}). 

In contrast, in GATv2, every node can attend to any other node using scoring function shown in eq.~\ref{GAT2}. 
\begin{equation}\label{GAT2}
    \overrightarrow h_i^{(k)} = \alpha_{ij}^{(k-1)} \Bigg[ f^{(k)} \Bigg(\mathbf{W}^{(k)} \cdot \sum_{j \in \mathcal{N}_i}  \overrightarrow h_j^{(k-1)} + \overrightarrow h_i^{(k-1)} \Bigg ) \Bigg]
    %{\color{red}\Bigg\vert~\cite{brody2021attentive}} 
\end{equation}

% \begin{gather}[right=quad\color{red} \empheqlVert]%
% 2A + 2B = 0\\
% A - B = 1,\\
% \sum_{i=0}^{A} = B\\
% \int_{0}^{A} x\,\mathrm{d}x = B
% \end{gather}
%We only inject the graph structure by only allowing node $i$ to attend over nodes in its neighborhood, $j\in \mathcal{N_i}$. These coefficients are then typically normalized using the softmax function, in order to be comparable across different neighborhoods:
%\begin{equation}\label{alphaij}
%\begin{aligned}
%    \alpha_{ij} & = \frac{\exp{\eta(\overrightarrow h_i,\overrightarrow h_j)}}{\sum_{k \in \mathcal{N}_i} \exp{\eta(\overrightarrow h_i,\overrightarrow h_k))}} \\
 %& = ReLU\Big(a^T \mathbf{W}[\overrightarrow {h_i} || \overrightarrow {h_j}]\Big)
%\end{aligned}
%\end{equation}
The main problem in the standard GAT scoring function (see eq.~\ref{vanillaGAT}) is that the learned layers $\mathbf{W}$ and $\alpha$ are applied consecutively, and thus can be collapsed into single linear layer~\cite{brody2021attentive}. To fix this limitation in our work, we then impose a relational inductive bias in data using neighborhood features aggregation (see eq.~\ref{proposedstagnn1} \&~\ref{proposedstagnn2}). In our proposed SAG-NN, the node i's embedding at step k for $k = 1$ is:
\begin{equation}\label{proposedstagnn1}
    \overrightarrow h_i^{(k)} = f^{(k)} \Bigg( \mathbf{W}^{(k)} . \Bigg[ AGG_{j \in \mathcal{N}_i}(\{\overrightarrow h_j^{(k-1)}\}), \overrightarrow h_i^{(k-1)}\Bigg] \Bigg),  
\end{equation}
where $\forall i\in V$ and $AGG(.)$ is the aggregation of i's neighbour's embeddings at step $k-1$ and $h_i^{(k-1)}$ is the node i's embedding at step $k-1$. And node i's embedding at step k for $k = 2, 3, \dots$ upto $K$ is:
\begin{equation}\label{proposedstagnn2}
    \overrightarrow h_i^{(k)} = f^{(k)} \Bigg(\mathbf{W}^{(k)} . \Bigg[\sum_{j \in \mathcal{N}_i} \alpha_{ij}^{(k-1)} \overrightarrow h_j^{(k-1)} + \alpha_{ii}^{(k-1)} \overrightarrow h_i^{(k-1)} \Bigg] \Bigg)
\end{equation}

The proposed solution not only improves the aggregation of features from neighbouring nodes, it also improves the ranking of attended nodes (static attention) as shown in eq. \ref{proposedstagnn1} \& \ref{proposedstagnn2}. 

\subsubsection{Spatio-temporal Classification via SAG-NN-E}\label{spatiotemporaltraining}
\begin{figure}[h!]
    \centering
    \includegraphics[scale=0.30]{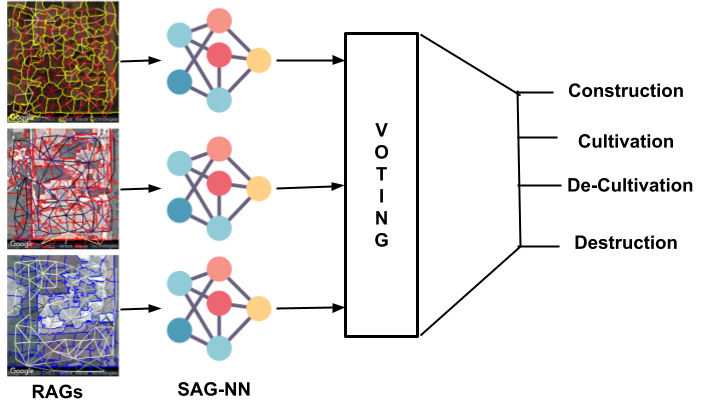}
    \caption{Spatio-temporal Classificcation via SAG-NN-E.}
    \label{fig:vaiSAG-NN}
\end{figure}

Although the proposed SAG-NN architecture is developed to account for neighborhood features' aggregation to learn spatial land-use classes, we also extended it for spatio-temporal classification. Given $T$ time steps, our resulting  ensemble SAG-NN-E has $T$ copies of SAG-NN, one for each time step, connected in parallel. The ensemble has a voting scheme that takes the spatial classification from each SAG-NN and generates the spatio-temporal classification (see Fig.~\ref{fig:vaiSAG-NN}). We used this ensemble as a baseline for evaluation of our proposed Spatio-temporal driven Graph Attention
Neural Network which is discussed next. % Then we chose a transition class having the most votes (e.g. if one SAG-NN classify RAG as barren land and other two classify as house, we say it's construction and vice versa) through voting mechanism.

%\subsubsection{Spatio-temporal Classificcation via SAG-NN}\label{spatiotemporaltraining}

%{\bf Semi-Deterministic}: We first sequentially classified the RAGs of images from different years from the C2D2 dataset, into one of the 14 land-use classes from the Asia14 dataset, i.e., farms, dense forests, houses, etc. Then in the second stage, we used a deterministic voting technique to label the sequence of land-use classification into one of the four key transitions. This approach gave us the baseline which we could compare to the previous work of Bhimra et al. \cite{bhimra2019using}. We used the SAG-NN architecture to account for neighborhood features' aggregation to learn spatial land-use classes. Then we chose a transition class having the most votes (e.g. if one SAG-NN classify RAG as barren land and other two classify as house, we say it's construction and vice versa) through voting mechanism.

\subsection{Spatio-temporal driven Graph Attention Neural Network with Block Adjacency matrix (STAG-NN-BA)}
\label{blockadjanceymatrixfortrags}
%\vspace{-0.2cm}
Images having multiple channels such as in case of color or multi-spectral images or sequence of multiple images are usually represented as a spatio-temporal volume. These patio-temporal volumes have fixed spatial dimension or pixels at each depth of the volume. However, when instead of pixels, superpixels of images are used this result in different dimension at each time step. Thus graph from superpixels of each image from a sequence cannot be stacked together as in case of pixel based representation. Furthermore, in GNNs the structure of the graph remains unchanged over multiple layers, only the node representation changes~\cite{kipf2016semi}. This restricts the use of GNNs for spatio-temporal classification problems having varying nodes over time. 

We addressed this problem by proposing a novel temporal-RAG that connects the individual RAG from each image. To incorporate the temporal change in graphs, we add the fourth dimension in the node features of these RAGs which is basically a numeric index that indicates the chronological order of the image the superpixel belongs to. We then combine the RAGs of these separate images into a supergraph that has these RAGs as unconnected subgraphs, we call this supergraph \emph{Temporal-RAGs}. Figure \ref{fig:TRAG} depicts the creation of \emph{Temporal-RAGs} from Images of a geo-location from different years. Our proposed temporal-RAG is an extension of our SAG-NN architecture. The supergraph of SAG-NN's is generated by combining the adjacency matrices from each RAG into a single adjacency matrix (see Figs.~\ref{fig:TRAG}). This results in a block diagonal adjacency matrix for Temporal-RAGs resulting in Spatio-temporal driven Graph Attention Neural Network with Block Adjacency matrix (STAG-NN-BA) defined as:

\begin{align}\label{eq:stag-nn4}
\begin{split}
\overrightarrow{h}_{i}^{(k)} & = \tiny ReLU \Bigg(\mathbf{W}^{(k)} . \Bigg[\sum_{j \in \mathcal{N}_i} \alpha_{ij}^{(k-1)} \overrightarrow h_j^{(k-1)} + \alpha_{ii}^{(k-1)} \overrightarrow h_i^{(k-1)} \Bigg] \Bigg) \\
& \mdoublepluss ReLU \Bigg(\mathbf{W}^{(k)} . \Bigg[\sum_{j \in \mathcal{N}_i} \alpha_{ij}^{(k-1)} \overrightarrow h_j^{(k-1)} + \alpha_{ii}^{(k-1)} \overrightarrow h_i^{(k-1)} \Bigg] \Bigg) \\
& \mdoublepluss ReLU \Bigg(\mathbf{W}^{(k)} . \Bigg[\sum_{j \in \mathcal{N}_i} \alpha_{ij}^{(k-1)} \overrightarrow h_j^{(k-1)} + \alpha_{ii}^{(k-1)} \overrightarrow h_i^{(k-1)} \Bigg] \Bigg)
\end{split}
\end{align}
where $\mdoublepluss$ symbol represent the concatenation of features.

\begin{comment}
$ \alpha_{ij} = ReLU \Bigg( \sum_{j\in \mathcal{N}_i} a^T\mathbf{WX_1} \Bigg) + ReLU \Bigg( \sum_{j\in \mathcal{N}_i} a^T\mathbf{WX_2} \Bigg)\\ + ReLU \Bigg( \sum_{j\in \mathcal{N}_i} a^T\mathbf{WX_3} \Bigg)~~ \textbf{STAG-NN-BA-v1 [Our]}\\$
\end{comment}
%%WadoodStart

In STAG-NN-BA we aggregate the node embedding from all the RAGs into one graph embedding {$X_G$} of length $D$. Then, we feed that embedding to the Multi-Layer Perceptron (MLP) for assigning one of the final transition classes. Our proposed architecture allows to impose relational inductive bias in data using neighborhood features aggregation over space as well as time resulting in a single architecture for data with a varying number of nodes over time (see Fig.~\ref{fig:stag-nn-ba12}).  Thus it can be used to classify the transitions or change in land-use over time in the remote sensing data. Since transitions are essentially temporal phenomena, the proposed STAG-NN-BA method can incorporate temporal information into regional adjacency graphs. We believe that this method can be extended to other geometric data. 

We do not assign features to the edges, since our model uses an attention mechanism, and we believe that the edge features will be learned according to the features of the connecting nodes. STAG-NN-BA combine ideas of graph convolutions \cite{kipf2016semi}, which allows graph nodes to aggregate information from their irregular neighbourhoods, with self-attention mechanisms \cite{vaswani2017attention}, which allows nodes to learn the relative importance of each neighbour during the aggregation process.  

Although, there are many different models that try to incorporate weights in neighborhood aggregation such as SplineCNN \cite{fey2018splinecnn} and GEO-GCN \cite{spurek2019geometric}. We used three approaches to perform a land-use transition classification of temporal images namely SAG-NN-E (see section~\ref{spatiotemporaltraining}), Global Sum Pooling (STAG-NN-BA-GSP) and Global Concatenated Pooling (STAG-NN-BA-GCP). The last two are discussed as follows:

{\bf Global Sum Pooling (STAG-NN-BA-GSP)}: There exist many different types of order-in-variant read-out layers in the literature, such as Global Average Pooling \cite{lin2013network}, Global Attention Pooling \cite{li2015gated}, Global Max Pooling \cite{lin2013network}, and Global Sum Pooling \cite{li2015gated}.

We use Global Sum Pooling (GSP) for it's simplicity as defined in the equation: $\mathbf{x}_{\mathcal{G}} = \sum_{v \in \mathcal{V}} \mathbf{x}^{(L)}_v$, where $V$ is the set of vertices, $\mathbf{x}^{(L)}_v$ is the node embedding at the last layer of a graph neural network, and $\mathbf{x}_{\mathcal{G}}$ is the embedding for the graph obtained as a result of the pooling operation. 

{\bf Global Concatenated Pooling (STAG-NN-BA-GCP)}:
We are using RAGs of images from three different timestamps combined into one Temporal-RAGs for the transition classification. Taking the graph readout in the last layers of GAT using Global Sum Pooling (GSP) adding all the nodes of the Temporal-RAGs into one $n$-dimensional vector. This makes the embedding of a Temporal-RAG indistinguishable from the embedding of a Temporal-RAG in which the underlying RAGs were to swap places. To solve this problem, we introduced a variation of GSP which gives us separate embedding for each underlying RAG concatenated into one $n\times D$ vector (see Fig.~\ref{fig:stag-nn-ba12}).

\begin{table*}
\centering
  \caption{Spatial classification accuracy on pixel based Region Adjacency Graph (RAG) of MNIST~\cite{lecun1998mnist} and subset of Asia14~\cite{nazir2020kiln} datasets. Top-2 ranking methods are in bold and, in particular, red (1st) and violet (2nd).}
  \scalebox{0.95}{
    \begin{tabular}{|c|c|c|c|}
    
    \hline
   {Architectures} &  {\#Param (M)} & {MNIST} & {Asia14}   \\ \hline
    \multicolumn{4}{|c|}{Classical models of neural network on image dataset}\\
    \hline
    Inception-ResNet-v2~\cite{szegedy2017inception}& 23.50 & - & 57.70 \% \\ \hline
    2D-ResNet-50~\cite{he2016deep}& 23.50 & - & 56.45 \% \\ \hline
    \multicolumn{4}{|c|}{Graph neural networks}\\
    \hline
    MoNet~\cite{monti2017geometric} & \textcolor{violet}{\bf 2.12}  & 91.11\% & 66.39\%  \\ \hline
    ChebNet~\cite{defferrard2016convolutional} & 12.85 & 75.62\% & 64.60 \% \\ \hline
    GATv1~\cite{velickovic2017graph} & 25.70 & 96.19\% & 69.85 \%   \\ 
    \hline
    AGNN~\cite{thekumparampil2018attention} & \textcolor{red}{\bf 0.41} & 97.98\% & 47.80\%   \\ 
    \hline
    %Dynamic GCN~\cite{yue2019dynamic}   &  & 68.60\%   \\ \hline
    GraphSAGE~\cite{hamilton2017inductive}  & 12.85  & 97.27\% & 70.00\%   \\ \hline
    Crystal GCN~\cite{xie2018crystal} & \textcolor{red}{\bf 0.41} & \textcolor{violet}{\bf 98.04\%} & 63.20\%   \\ \hline
    %CGCNN~\cite{xie2018crystal}  &  &53.80\%    \\ \hline
    %96.20\%  & 85.93\% \\
    
    GATv2 \cite{brody2021attentive}& 25.70 & - & \textcolor{violet}{\bf 71.10\%} \\ \hline
   % Transformer \cite{vaswani2017attention}& -& - & \textcolor{red}{{\bf 77.10\%}} \\ \hline   
     SAG-NN (our) & 25.69  & \textcolor{red}{\bf 98.14\%} & \textcolor{red}{\bf 77.00\%}  \\
    \hline
    \end{tabular}}
  \label{tab:pixelbased}%
\end{table*}%

\begin{table*}
\centering
  \caption{Spatial classification accuracy on SLIC superpixels based Region Adjacency Graph (RAG) of subset of Asia14~\cite{nazir2020kiln} datasets. Top-2 ranking methods are in bold and, in particular, red (1st) and violet (2nd).}
  \scalebox{0.95}{
    \begin{tabular}{|c|c|c|}
    
    \hline
   {Architectures} &  {\#Param (M)} & {Asia14}   \\ \hline
    \multicolumn{3}{|c|}{Classical models of neural network on image dataset}\\
    \hline
    Inception-ResNet-v2~\cite{szegedy2017inception}& 23.50  & 57.70 \% \\ \hline
    2D-ResNet-50~\cite{he2016deep}& 23.50  & 56.45 \% \\ \hline
    \multicolumn{3}{|c|}{Graph neural networks}\\
    \hline
    GCN~\cite{kipf2016semi} & \textcolor{red}{\bf 0.015} &  9.78\%   \\ 
    \hline
    GraphSAGE~\cite{hamilton2017inductive} & \textcolor{red}{\bf 0.015} & 65.00\% \\ \hline
    GATv1~\cite{velickovic2017graph} & \textcolor{violet}{\bf 0.030} & \textcolor{violet}{\bf 80.30} \\  \hline
    GATv2~\cite{brody2021attentive} & 0.055 & 72.04\% \\ \hline
    SAG-NN (our) & \textcolor{violet}{\bf 0.030}  & \textcolor{red}{\bf 80.98\%}  \\
    \hline
    \end{tabular}}
  \label{tab:superpixelbased}%
\end{table*}%
\section{Results and Evaluation}
%We evaluated the proposed approach and SOTA GNN classifiers on pixel-based and superpixel based graph representation of images. The datasets, training parameters, quantitative and qualitative evaluation are discussed in following sections.

\subsection{Datasets}
We used three datasets for evaluation of our proposed approch namely MNIST~\cite{lecun1998gradient}, Asia14~\cite{nazir2020kiln} and C2D2 Dataset~\cite{bhimra2019using}. Both Asia14 and C2D2 datasets are remote sensing datasets for spatial and spatio-temporal classification respectively. These datasets capture graph signal classification tasks, where graphs are represented in mixed mode: one adjacency matrix, and many instances of node features. Details of these datasets are discussed next. 

\subsubsection{MNIST Pixel-based Dataset}
The MNIST dataset~\cite{lecun1998gradient} is an acronym that stands for the Modified National Institute of Standards and Technology dataset. It is a dataset of $28\times28$ pixel grayscale images of handwritten single digits between $0$ and $9$. MNIST dataset containing $70,000$ pixel based region adjacency graphs as described by \cite{defferrard2016convolutional}. Every graph is labeled by one of 10 classes.

\subsubsection{Asia14 pixel-based and Superpixels Dataset}
%Unlike existing methods~\cite{boyd2018slavery, misrabrick} that analyze only one specific region, 
\emph{Asia14} dataset contains samples under varying conditions as discussed in Section~\ref{sec:chsol}. Furthermore, unlike street imagery, land-use is subject to significant variations in satellite imagery. To cater for this, we used a subset of $14$-class dataset named \emph{Asia14}~\cite{nazir2020kiln}. This dataset consisting of Digital Globe RGB band images from 2016 and 2017 of resolution $256\times 256$ at zoom level $20$ (corresponding to $0.149$ pixel per meter on the equator). We used $9$ classes including brick kilns, houses, roads, tennis courts, grass, dense forest, parking lots, parks. The issue of sensor variations is handled by diversifying the training data across several spatial locations within the Indo-Pak region of South Asia. There are $9,000$ pixel-based region adjacency graphs and we generated the superpixels using SLIC~\cite{achanta2012slic}. Then $9,000$ graphs, with $75$ nodes each, were generated using region adjacency graph method.

\subsubsection{C2D2 Dataset}
This dataset contains Spatio-temporal data annotated for four fundamental land-use land-change transitions namely construction, destruction, cultivation, and de-cultivation. This dataset was originally collected and prepared by \cite{bhimra2019using}. They browsed Digital Globe imagery data for the years 2011, 2013, and 2017 and visited almost $5,50,000$ random locations which make approximately 5310 $km^2$. Along with lat-long, at each location, we cropped an image patch of  resolution $256\times 256$ at zoom level $20$ (i.e. $0.149$ pixel per meter on the equator). The provided dataset contained 3D volumes of Spatio-temporal images from different years. we had to reverse the process to separate out the individual images for a location into the directories of each year. We then generate regional adjacency graphs (RAG)s from the superpixels of these images that were generated using SLIC and use the same annotations as it was assigned to the 3D volumes. 
%https://captain-whu.github.io/AID/

%\subsection{Hyperparameters for the training of all networks}
%\subsection{Quantitative Evaluation}
%We proposed two main approaches for the classification of the spatial and spatio-temporal land-use transition in geo-spatial images. In this section, we will be discussing the experiments and results we obtained from each methodology separately along with the discussion and comparison of obtained results with previous works. For the sake of better clarity, we briefly describe the pre-processing we did on the datasets of images to generate the corresponding graphs that are used in our experiments. The methods to represent images as graphs are reviewed in section~\ref{cdpre}. We have kept the same train, validation and test splits for all the models, i.e. $70\%$, $15\%$, and $15\%$ respectively. 

\begin{table*}[!h]
\caption{Spatio-temporal comparative evaluation for land-use transition classification on C2D2 dataset respectively. (Key: Acc.: Accuracy, Par.: Parameters, M: Millions, FPT: Forward Pass Time in milliseconds for 100 forward passes). Top-2 ranking methods are in bold and, in particular, red (1st) and violet (2nd).}
\label{Table:c2d2_results}
  \centering
  \scalebox{1}{
    \begin{tabular}{|c|c|c|c|}
    \hline
    Model & \# Par. (M) & FPT (ms) & Acc. \\
    \hline
    %\textbf{GATv2+Voting (ours)}  & 0.03 M & \textbf{3.6 ms}  & \textbf{60 \%} \\
    %\textbf{ST-GATv1~(1x Read-Out) (ours)}  & 0.03 M & 2.77 ms  & 48 \% \\
    3D-ResNet-34~\cite{bhimra2019using}& 63.50  & $>$ 3.6 & {57.72 \%} \\ \hline
    %Graph CNN~\cite{censi2021attentive} & - & - & 50 \%  \\
    %SAG-NN + voting (ours) & 0.030 & - & {\bf 82 \%} \\
    SAG-NN-E &  \textcolor{red}{\bf 0.030} & 3.60 ms  & 60.02 \% \\ \hline
    STAG-NN-BA-GCP (ours) & \textcolor{violet}{\bf 0.050}  & \textcolor{red}{\bf 2.50 ms}  & \textcolor{violet}{\bf 64.90 \%} \\ \hline
    STAG-NN-BA-GSP (ours) & \textcolor{red}{\bf 0.030}  & \textcolor{violet}{\bf 2.62 ms}  & \textcolor{red}{\bf 77.83 \%} \\

    \hline
    \end{tabular}}
 \end{table*}

\subsection{Evaluation of SAG-NN}
We evaluated our Spatial Attention Graph Attention Network (SAG-NN) architecture on two datasets namely MNIST and Asia14. We performed two experiments, in the first experiment we generated pixel-based graphs and in the second experiment we used superpixel based graphs. We performed comparisons with two classical methods namely Inception-ResNet-v2~\cite{szegedy2017inception} and 2D-ResNet-50~\cite{he2016deep} and seven graph based state-of-the-art methods namely MoNet~\cite{monti2017geometric}, ChebNet~\cite{defferrard2016convolutional}, GATv1~\cite{velickovic2017graph} , AGNN~\cite{thekumparampil2018attention}, GraphSAGE~\cite{hamilton2017inductive}, Crystal GCN~\cite{xie2018crystal}, GATv2 \cite{brody2021attentive}. 

We first trained and validated our Spatial Attention Graph Attention as well as all the other methods on MNIST dataset. Our SAG-NN model achieved highest accuracy of $98.14\%$ on MNIST dataset with $25.69$ million number of parameters on pixel-based RAG. Then we trained and tested SAG-NN  as well as all the other methods on Asia14 dataset. Here again our proposed SAG-NN achieved highest test accuracy of $77.00\%$ and $80.98\%$ on pixel-based graph and superpixel RAGs respectively (see Table~\ref{tab:pixelbased} and \ref{tab:superpixelbased}). %The training parameters for superpixel RAGs are $0.030$ million only. SAG-NN performs better than all SOTA graph neural network with comparable low compute cost.

In Table~\ref{tab:pixelbased}, the experiments show that the SAG-NN outperforms on pixel-based RAGs as compared to other classical or RAG-based GNN classifiers. In Table~\ref{tab:superpixelbased}, SAG-NN has comparable training parameters and shows high accuracy when compared with GCN~\cite{kipf2016semi} and GraphSAGE~\cite{hamilton2017inductive}. It shows comparable high accuracy when compared with GATv1~\cite{velickovic2017graph}. GATv2~\cite{brody2021attentive} is proposed for bipartite graphs that's why it shows low performance on pixel-based and superpixel-based region adjacency graphs as compared to our proposed model.

\subsection{Evaluation of STAG-NN-BA}
 We compared both the variants of our STAG-NN-BA with two other methods namely 3D-ResNet-34~\cite{bhimra2019using} and SAG-NN-E. SAG-NN-E is our extension of SAG-NN for spatio-temporal data and serves as the baseline. 3D-ResNet-34~\cite{bhimra2019using} on the other hand uses 3D convolution and is the only state-of-the-art method with published results on C2D2 dataset. In order to compare our results on C2D2 dataset, we used the same train/test split as in 3D-ResNet-34~\cite{bhimra2019using}. The ability of transition classification for SAG-NN-E approach is dependent on the performance of land-use classification and voting procedure (see section~\ref{spatiotemporaltraining}). Both STAG-NN-BA-GCP and STAG-NN-BA-GSP achieved significantly higher accuracies as compared to SAG-NN-E and 3D-ResNet-34~\cite{bhimra2019using} in terms of accuracy and compute cost. STAG-NN-BA-GCP and STAG-NN-BA-GSP achieved approximately $7\%$ and $20\%$ higher accuracy as compared to 3D-ResNet-34. They also achieved $4.88\%$ and $17.81\%$ higher accuracy as compared to SAG-NN-E which indicates the effectiveness of our temporal model STAG-NN-BA as compared to spatial model via SAG-NN. Furthermore, STAG-NN-BA-GSP ourperforms all the other methods which shows that the global sum pooling is a more suited method of aggregation as compared to global concatenated pooling.
 
 Table~\ref{Table:c2d2_results} also compares the training parameters, forward pass time, and accuracy of our models used for spatio-temporal land-use classification. It can be seen that the forward pass time of STAG-NN-BA is almost $1$ms lower as compared to SAG-NN and much lower as compared to 3D-ResNet-34.
%The frame by frame inference on 2D images with voting using architectures like Inception-ResNet-v2, 2D-ResNet-50 is compared with our similar model for graph by graph inference of RAGs using GATv1 and voting. 
%Additionally, we compared our novel Spatio-Temporal-GATv1 (ST-GATv1) on Temporal-RAGs with 3D-ResNet-34 architecture purposed by Bhimra et al. As shown in the table, our semi-deterministic model achieved better accuracy then every other approach. We attribute this performance to the high accuracy of GATv1 that was trained on Asia14 dataset. 

In the land-use transition classification, the STAG-NN-BA-GSP approach is the most reliable. However, we also draw comparison of 3D-ResNet-34~\cite{bhimra2019using} with SAG-NN-E and STAG-NN-BA-GCP (see Table~\ref{Table:c2d2_results}). Both spatio-temporal proposed models (STAG-NN-BA-GCP and STAG-NN-BA-GSP) achieved higher performance with low computational cost on the C2D2 dataset. 

%and the complexity of the model is reduced exponentially. Obviously, the train time and the forward pass time (FPT) is expected to be decreased with similar magnitudes.

%Forward pass time (fpt)

% \begin{figure}[h!]
%   \centering
% \includegraphics[scale=0.178]{Figures/results/qualitative_example.png}
% \caption{Classification for a Spatio-temporal example by different model for qualitative comparison. SAG-NN+voting classifies it as \emph{Cultivation} and STAG-NN-BA-GSP classifies it as \emph{Construction}.}
% \label{fig:results_qualt}
% \end{figure}

\section{Conclusion and Future work}
This paper proposed two novel Graph Neural Network architectures for spatial and spatio-temporal classification of remote sensing imagery to gain a deeper understanding of land-use and by extension socio-economic indicators. We also proposed a novel method to represent temporal information in images using region adjacency graph called Temporal-RAG. We evaluated our approaches on two remote sensing datasets namely Asia14 and C2D2. The comparison with the previously existing classical and graph neural network methods showed that our approaches achieved higher performance and reduced the computation power greatly. There are two areas recognized while working on this paper that can serve as interesting problems for future works. Firstly, there is an issue of information loss during the generation of graphs from superpixel segmentation. Secondly, over-segmentation of an image to make superpixels causes information loss, which decreases the representation power of pixels-based graphs. %Average color values of all the pixels of image that belong to a superpixel segment are the only meaningful features for graph node based on that superpixel. 
The information about the shape of the underlying superpixel segment is lost. We can extract generic shape embedding using an auto-encoder into a single $N$ dimensional vector. While assigning the color values as features, this $N$-dimensional shape embedding vector can be concatenated into the initial features. This can help incorporate the shape into graph representations.

%Since there are no edges between nodes for different RAGs, this impedes the information flow while message passing in Temporal-RAGs. We can employ different techniques such as for connecting the nodes of underlying RAGs to make a connected graph. 
%A list of some promising metrics for connecting RAGs is KL Divergence, Intersection Over Union (IOU), Edge Prediction, etc.

%\clearpage\mbox{}Page \thepage\ of the manuscript.
%\clearpage\mbox{}Page \thepage\ of the manuscript.

%This is the last page of the manuscript.
%\par\vfill\par
%Now we have reached the maximum size of the ECCV 2020 submission (excluding references).
%References should start immediately after the main text, but can continue on p.15 if needed.

%\clearpage

% \bibliographystyle{ACM-Reference-Format}
\bibliography{egbib}

\begin{comment}
$(see Table~\ref{tab:socioeconomic})$
\begin{table*}[!h]
\centering
  \caption{Classifications of Proposals for Predicting Socio-Economic Indicators}
  \scalebox{0.52}{
    \begin{tabular}{|c|p{3.4cm}|p{2.7cm}|c|}
    \hline
   \bf{Proposals} & \bf{Indicators} & \bf{Prediction} & \bf{Method} \\ \hline
    Censi A. M, et al.~\cite{censi2021attentive}& Land Cover Mapping & Earth Surface Dynamics & Graph CNN \\ \hline
    Nazir, Usman, et al. (2019)~\cite{nazir2019tiny} & Brick Kilns & Slavery & CNN \\ \hline
    Bhimra, M. Ahmed, et al. (2019)~\cite{bhimra2019using} & Barren Land, Trees, Buildings, Crops & Construction, Destruction, Cultivation, Decultivation  & 3D CNN \\ \hline
    Piaggesi, Simone, et al. (2019)~\cite{piaggesi2019predicting} & Household Income & Poverty & Transfer Learning \\ \hline 
    Boyd, Doreen S., et al. (2018)~\cite{boyd2018slavery}& Brick Kilns & Slavery & Manual Annotation \\ \hline
    You, Jiaxuan, et al. (2017)~\cite{you2017deep}& Soybeans Cropland & Crop Yield & LSTM + CNN + Gaussian Process \\ \hline  
    Hordiiuk, D. M., et al. (2017)~\cite{hordiiuk2017neural}  & Buildings & Urban Damage Detection & CNN + Laplace Filter \\ \hline
    Xie, Michael, et al. (2016)~\cite{xie2016transfer}  & Nighttime Light Intensity & Poverty& Transfer Learning \\ \hline
\end{tabular}}%
  \label{tab:socioeconomic}%
\end{table*}%
\end{comment}
%\textcolor{red}{THE FIRST PARA IS TOO HEAVY ON IMAGENET, WE SHOULD DILUTE IT AND MIX IT WITH REMOTE SENSING. THIS WILL MAKE IT CONSISTENT WITH LAST PARA OF THIS SECTION}

\newpage

%\appendix

\section{Supplementary Material}

\begin{appendices}

\begin{figure}[!h]
\centering
\begin{tabular}{ccc}
        \includegraphics[ width=.25\columnwidth]{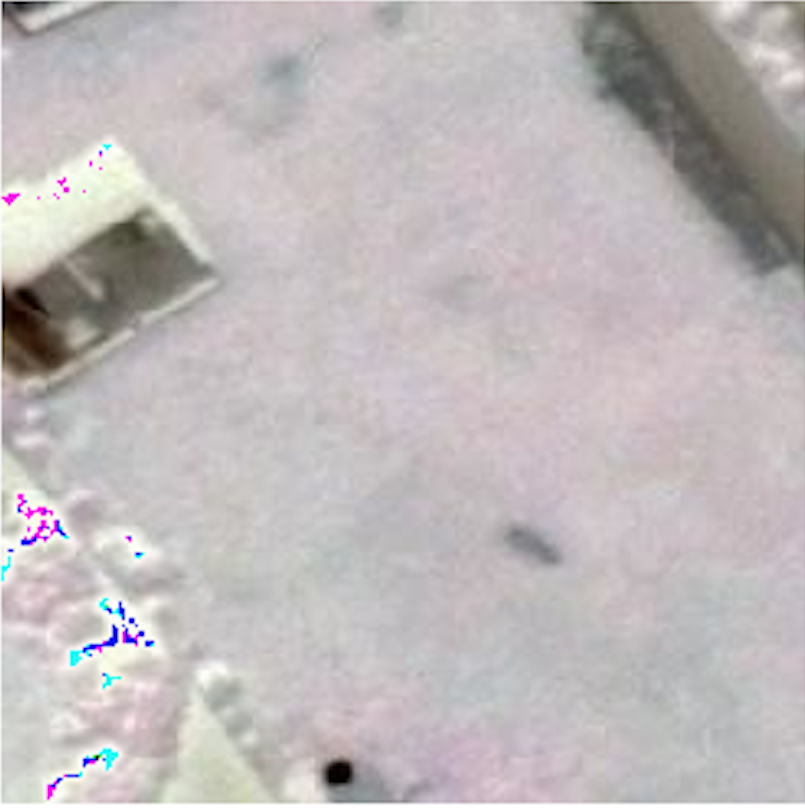} &\includegraphics[ width=.25\columnwidth]{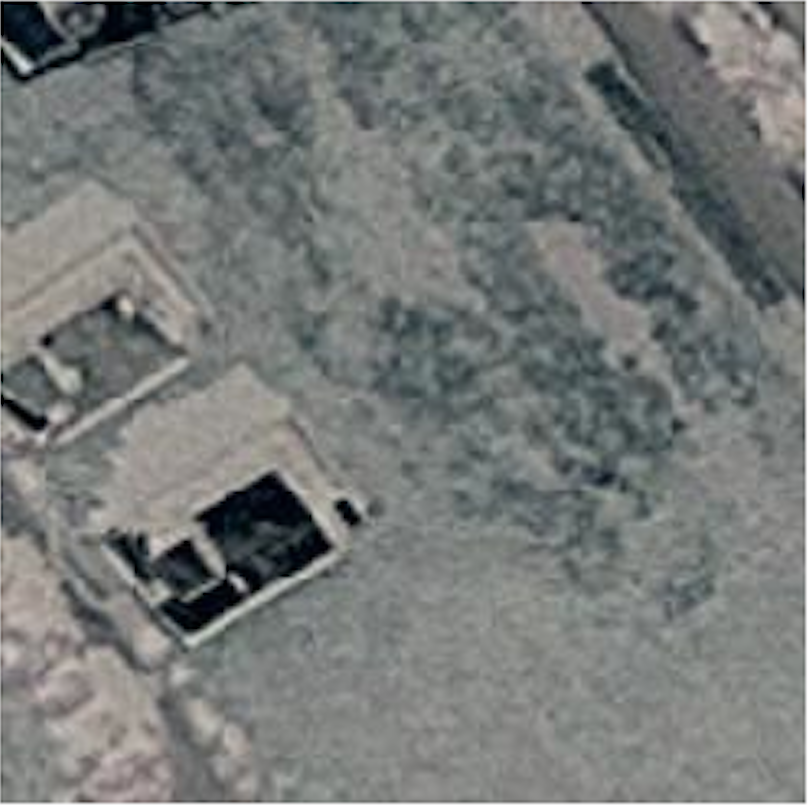}&
		\includegraphics[ width=.25\columnwidth]{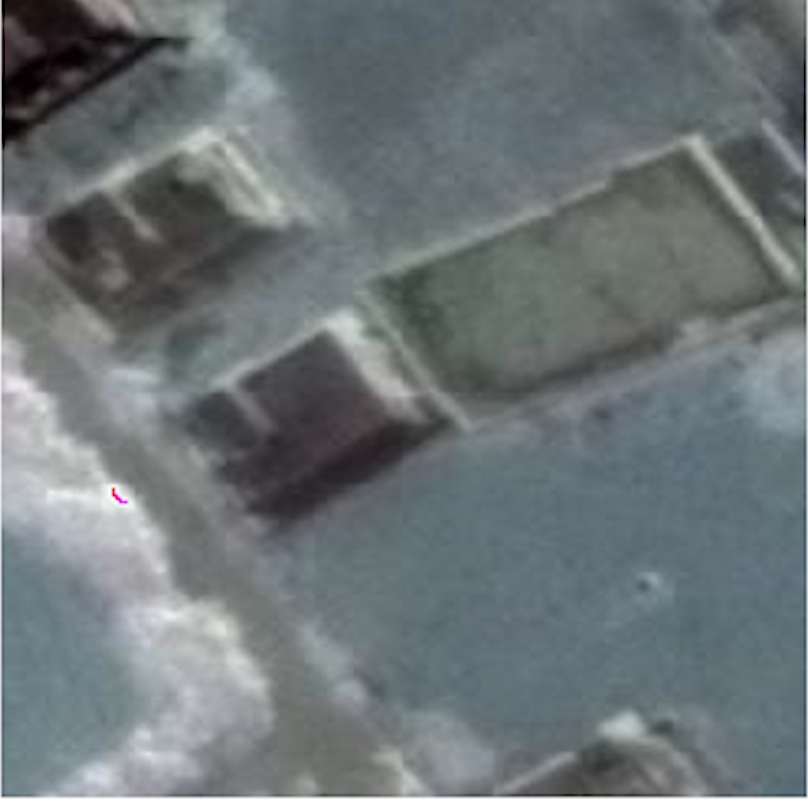}\\
 		\includegraphics[ width=.25\columnwidth]{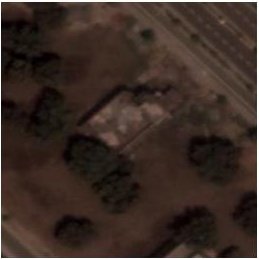} &\includegraphics[ width=.25\columnwidth]{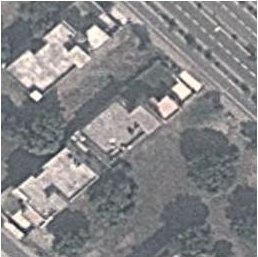}&
		\includegraphics[ width=.25\columnwidth]{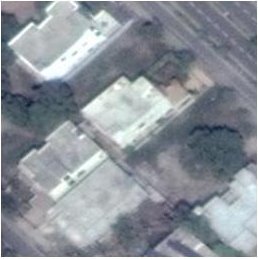}\\
		\includegraphics[ width=.25\columnwidth]{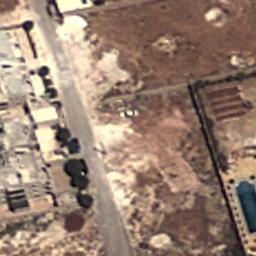} &\includegraphics[ width=.25\columnwidth]{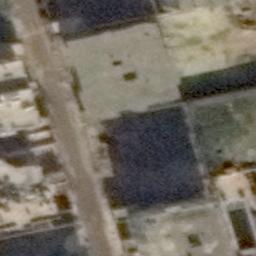}&
		\includegraphics[width=.25\columnwidth]{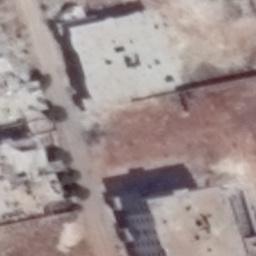}\\
		2011 & 2013 & 2017 \\ 
	\end{tabular}
	\caption{Examples showing the change in land-use between 2011 and 2017. In all three examples, more and more land was used for construction purposes over the years. See Section \ref{ssec:QA} for a discussion on results. (Satellite images courtesy Google Earth).}
	\label{QR}
\end{figure}
\section{Implementation Details}
All the graph neural networks are trained using PyTorch. Optimization method is Adam with an initial learning rate of $ 1e^{-3}$. The learning rate increases by $0.1$ if validation loss does not decline for $20$ epochs. Instead of using fixed number of epochs, we used early stopping criteria and patience for early stopping is $200$. We have kept the same train, validation and test splits for all the datasets, i.e. $70\%$, $15\%$, and $15\%$ respectively. 
%The regularization rate for l2 is $5e^{-4}$.

\section{Qualitative Analysis}

\label{ssec:QA}
Fig. \ref{QR} shows the sample annotations for \emph{Construction} transition class. In Fig. \ref{QR} (Row 1), SAG-NN-E with voting mechanism classifies it as \emph{Cultivation} which is clearly wrong as it can be seen from the middle and last image that the land has undergone the \emph{Construction}. This type of misclassification is expected from the model since there are two transitions in three images of geolocation. The voting mechanism tends to get confused when multiple transitions are present in an example. But our proposed model `STAG-NN-BA-GSP' correctly classifies it as \emph{Construction}. In Fig. \ref{QR} (Row 2) our all models: SAG-NN-E, STAG-NN-BA-GCP, and STAG-NN-BA-GSP classify it as \emph{Construction}. In Fig. \ref{QR} (Row 3), SAG-NN-E and STAG-NN-BA-GSP correctly classify it but STAG-NN-BA-GCP confused it with \emph{Destruction} perhaps because in this example one building is removed while multiple others were added. 
\end{appendices}

\end{document}